\documentclass[conference]{IEEEtran}
\IEEEoverridecommandlockouts
\usepackage{cite}
\usepackage{amsmath,amssymb,amsfonts}
\usepackage{graphicx}
\usepackage{textcomp}
\usepackage{xcolor}
\usepackage{multirow}
\usepackage{algorithm,algpseudocode}
\def\BibTeX{{\rm B\kern-.05em{\sc i\kern-.025em b}\kern-.08em
    T\kern-.1667em\lower.7ex\hbox{E}\kern-.125emX}}

\makeatletter
\newcommand{\linebreakand}{%
\end{@IEEEauthorhalign}
\hfill\mbox{}\par
\mbox{}\hfill\begin{@IEEEauthorhalign}
}
\makeatother

\begin{document}

\title{HMS-OS: Improving the Human Mental Search Optimisation Algorithm by Grouping in both Search and Objective Space\thanks{This research is funded within the SFEDU development program (PRIORITY 2030).}
}

\author{
\IEEEauthorblockN{Seyed Jalaleddin Mousavirad}
\IEEEauthorblockA{\textit{Computer Engineering Department} \\
\textit{Hakim Sabzevari University}\\
Sabzevar, Iran }
\and
\IEEEauthorblockN{Gerald Schaefer}
\IEEEauthorblockA{\textit{Department of Computer Science} \\
\textit{Loughborough University}\\
Loughborough, U.K.}
\and
\IEEEauthorblockN{Iakov Korovin}
\IEEEauthorblockA{\textit{Southern Federal University} \\
	Taganrog, Russia\\ }
\linebreakand
\IEEEauthorblockN{Diego Oliva}
\IEEEauthorblockA{\textit{Depto. de Ciencias Computacionales} \\
	\textit{Universidad de Guadalajara}\\
	Guadalajara, Mexico\\ }
\and
\IEEEauthorblockN{ Mahshid Helali Moghadam}
\IEEEauthorblockA{\textit{RISE Research Institutes of Sweden} \\
	\textit{M\"alardalen University}\\
	V\"aster\'as, Sweden\\}
\and
\IEEEauthorblockN{Mehrdad Saadatmand}
\IEEEauthorblockA{\textit{RISE Research Institutes of Sweden} \\
	Sweden\\}
}

\maketitle

\begin{abstract}
The human mental search (HMS) algorithm is a relatively recent population-based metaheuristic algorithm, which has shown competitive performance in solving complex optimisation problems. It is based on three main operators: mental search, grouping, and movement. In the original HMS algorithm, a clustering algorithm is used to group the current population in order to identify a promising region in search space, while candidate solutions then move towards the best candidate solution in the promising region. In this paper, we propose a novel HMS algorithm, HMS-OS, which is based on clustering in both objective and search space, where clustering in objective space finds a set of best candidate solutions whose centroid is then also used in updating the population. For further improvement, HMS-OS benefits from an adaptive selection of the number of mental processes in the mental search operator. Experimental results on CEC-2017 benchmark functions with dimensionalities of 50 and 100, and in comparison to other optimisation algorithms, indicate that HMS-OS yields excellent performance, superior to those of other methods.
\end{abstract}


\section{Introduction}
Global optimisation algorithms aim to find the best among all feasible solutions for a given problem. Conventional algorithms such as gradient-based approaches are popular yet ineffective and sometimes impractical~\cite{Classic_ineefective01}, and also suffer from some difficulties such as their tendency towards local optima and requirements to calculate derivative information. Population-based metaheuristic algorithms, such as particle swarm optimisation~\cite{PSO_Main_Paper02}, can provide a reliable alternative. 

Population-based metaheuristic algorithms generally start with a set of randomly-generated candidate solutions (the population). They then employ some operators on the population to share information among different candidate solutions, and consequently, to move them towards a promising area in search space. According to the no free lunch (NFL) theorem~\cite{No_Free_Lunch} , there is no single best-suited algorithm for all optimisation problems, which has motivated significant research to develop effective population-based algorithms, both established~\cite{DE_Competition_ICCSE2019,DE_Deceptive_ICCSE2019,center_sampling_PSO_SMC2020} and more recent approaches~\cite{GWO_Main_Paper, WOA_Main_Paper,SCA_Main_Paper}.   

Human mental search (HMS)~\cite{HMS_Main_Paper,HMS_Main_Paper2} is a relatively recent population-based metaheuristic that has demonstrated competitive performance in solving complex optimisation problems with different characteristics~\cite{HMS_Main_Paper}. HMS has also been shown to outperform other algorithms in applications such as static and dynamic clustering-based image segmentation~\cite{HMS_Image_Clustering,HMS_Automatic_clustering,HMSI_Static_conf}, colour quantisation~\cite{HMSI_Quantisation,HMS_Quantisation_2}, and image thresholding~\cite{HMS_Image_Thresholding}, while also having been used for classical engineering problems such as welded beam design~\cite{HMS_Main_Paper}.   

Similar to other population-based algorithms, HMS starts with a population of candidate solutions (called bids in HMS) and is founded on three main operators: mental search, which explores the vicinity of each bid based on a Levy flight distribution; grouping, which identifies a promising region, and movement, which moves candidate solutions towards the promising region.

The grouping operator in HMS plays a crucial role. Here, a clustering algorithm ($k$-means in standard HMS) is applied on the current population. Each generated cluster then represents a region in search space, and the cluster with the best mean objective function value is considered as the promising region used in the subsequent movement operator. It is worthwhile to note that the promising region may not contain the best candidate solution of the current population.

\cite{Mid_Point} shows that the population mid-point (centre) can improve the efficacy of the population-based algorithms. The population mid-point not only converges to a local optimum but also enhances the efficacy of an algorithm in its exploitation phase, so that improved population-based algorithms can be devised using this mid-point strategy~\cite{center_sampling_PSO_SMC2020,center_DE_Mine01,DE_Deceptive_ICCSE2019}.

In this paper, we propose a novel HMS algorithm, HMS-OS, which is based on clustering in both objective and search space. Clustering in objective space partitions the population based on the objective function, so that the best area includes the best individuals. The movement operator then uses both the best individual of the promising region and the average (mid-point) of the best objective space cluster. For further improvement, HMS-OS employs an adaptive mental search approach, inspired by~\cite{HMS-IS-OSK}, to adjust the number of mental searches used for each candidate solution.

The remainder of the paper is organised as follows. Section~\ref{Sec:HMS} describes the standard HMS algorithm, while Section~\ref{Sec:Proposed} introduces our novel HMS-OS approach. Section~\ref{Sec:Exp} provides experimental results, and Section~\ref{Sec:conc} concludes the paper.

\section{Human Mental Search}
\label{Sec:HMS}
Human mental search (HMS)~\cite{HMS_Main_Paper,HMS_Main_Paper2} is a relatively recent population-based metaheuristic, which is inspired by the manner of a human operating in the bid space of online auctions. Each candidate solution in HMS is called a bid (or offer). HMS begins with a set of randomly-generated bids and is founded on three main operators: mental search, grouping, and movement, which are briefly explained in the following, while the overall HMS algorithm is given in form of pseudo-code in Algorithm~\ref{alg:HMS} (for a minimisation problem).

\begin{algorithm}[t!]
	\caption{Pseudo-code of HMS algorithm.}
	\label{alg:HMS}
	\begin{algorithmic}[1]
		\State // $L$/$U$: lower/upper bound; $M_{l}$/$M_{h}$: min./max.\ number of mental processes; $N_{pop}$: number of bids; $K$: number of clusters; $NFE_{\max}$:  max.\ number of function evaluations
		\State \
		\State Initialise population of $N_{pop}$ bids
		\State Calculate objective function values (OFVs) of all bids
		\State $x^{*}$ = best bid in the initial population
		\State $NFE=N_{pop}$
		\State $iter=0$
		\While {$NFE<=NFE_{\max}$}
		\State $iter=iter+1$
		\State // Mental Search
		\For {$i$ from 1 to $N_{pop}$}
		\State $\beta_{i}$ = random number between $L$ and $U$
		\State $q_{i}$ = integer random number between $M_{l}$ and $M_{h}$
		\For {$j$ from 1 to $q_{i}$}
		\State $S=(2-NFE\frac{2}{NFE_{\max}})0.01\frac{u}{v^{1/\beta _{i} } } (x_{i} -x^{*})$
		\State $NS_{j} = x_{i}+S$
		\EndFor
		\State $t$ = $NS$ with lowest OFV
		\If {OFV of $t$ $<$ OFV of $x_{i}$}
		\State $x_{i} = t$
		\EndIf
		\State $NFE=NFE + q_{i}$
		\EndFor
		
		\State // Grouping
		\State Cluster $N_{pop}$ bids into $K$ clusters
		\State Calculate mean OFV of each cluster
		\State winner cluster = cluster with lowest mean OFV 
		\State $W$ = best bid in winner cluster
		
		\State // Movement
		\For {$i$ from 1 to $N_{pop}$}
		\State $x_{i} = x_{i} +C(r W - x_{i})$
		\EndFor
		\State Calculate OFVs of all bids
		\State  $x^{+}$ = best bid in current bids
		\If {OFV of $x^{+}$ $<$ OFV of $x^{*}$}
		\State $x^{*} =x^{+}$
		\EndIf
		\State $NFE = NFE + N_{pop}$
		\EndWhile
	\end{algorithmic}
\end{algorithm}

\subsection{Mental Search}
The mental search operator explores the vicinity of each bid based on a Levy flight distribution where several small steps are mixed with sudden long jumps. Mental search thus improves both exploration and exploitation simultaneously. 

A new bid in the neighborhood of an existing bid $x_{i}$ is generated as
\begin{equation}
\label{eq:MS}
NS = x_{i} + S ,
\end{equation}
with 
\begin{equation}
\label{eq:Levy}
S = (2-NFE\frac{2}{NFE_{\max}}) 0.01 \frac{u}{v^{1/\beta } } (x_{i} -x^{*}) ,
\end{equation}
where $NFE$ is the number of function evaluations so far, $NFE_{\max}$ is the maximum number of function evaluations, $x_{i}$ indicates the current bid, $x^{*}$ is the best bid found so far, and $u$ and $v$ are random numbers calculated as
\begin{equation}
u\sim N(0,\sigma _{u}^{2} ) , \ \ v\sim N(0,\sigma _{v}^{2} ) ,
\end{equation}
with
\begin{equation}
\sigma _{u} =\left\{\frac{\Gamma (1+\beta )\sin (\frac{\pi \beta }{2} )}{\Gamma [(\frac{1+\beta }{2} )]\beta 2^{(\beta -1)/2} } \right\}^{1/\beta } ,\ \ \sigma _{v} =1 ,
\end{equation}
where $\Gamma$ is a standard gamma function.

\subsection{Grouping}
The goal of the grouping operator is to partition the current population in search space based on a clustering algorithm such as $k$-means. For each generated cluster, the mean objective function value is calculated, and the cluster with the lowest mean objective value (in a minimisation problem) identified as the winner cluster indicating a promising area in search space.

\subsection{Movement}
During movement, candidate solutions move towards $W$, the best bid of the winner cluster, as  
\begin{equation}
\label{eq:bidupd}
x_{i} = x_{i} + C (r  W - x_{i}),
\end{equation}
where $C$ is a constant, and $r$ is a random number in $[0,1]$ taken from the uniform distribution.

\section{HMS-OS Algorithm}
\label{Sec:Proposed}
In this paper, we improve the HMS algorithm through two modifications. The main contribution is to cluster the population in terms of both search space and objective space. The second one is to adaptively select the number of mental searches.

\subsection{Clustering in objective space }
\begin{figure}[b!]
	\centering
	\includegraphics[width=.95\columnwidth]{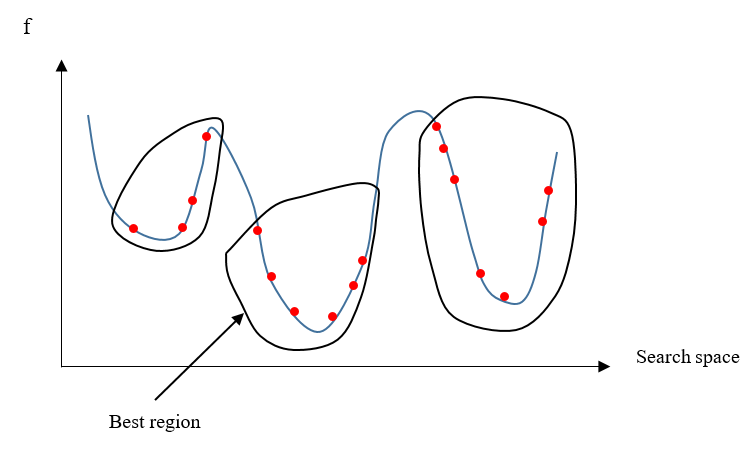}
	\caption{Population clustering in search space.}
	\label{fig:searchspace}
\end{figure}

Grouping in HMS is performed, using the $k$-means algorithm, in search space, i.e.\ the space by which candidate solutions are defined. This is illustrated in Fig.~\ref{fig:searchspace} for a sample population of 16 bids which get divided into three clusters. As can be observed, each cluster contains bids that are close to each other in search space, and we can also identify the cluster that represents the best region based on each cluster's mean objective function value. 

\begin{figure}[t!]
	\centering
	\includegraphics[width=0.95\columnwidth]{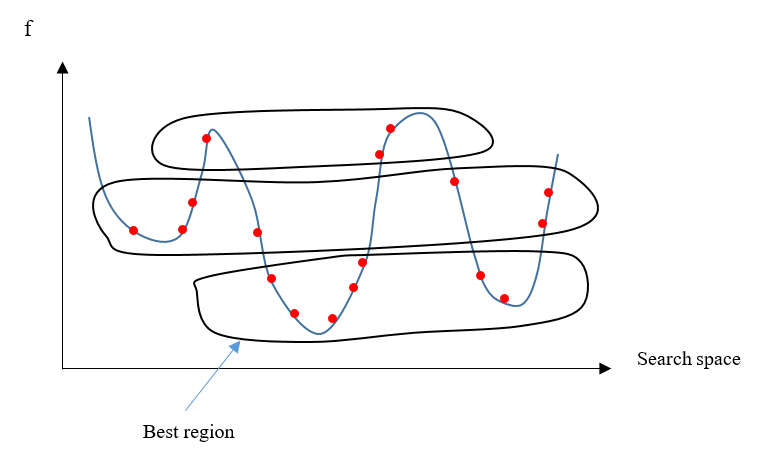}
	\caption{Population clustering in objective space.}
	\label{fig:objspace}
\end{figure}

In this paper, we propose to also cluster the population based on the (one-dimensional) objective space. That is, we use $k$-means to cluster based on the objective function values of candidate solutions. After clustering, the best region here contains the best candidate solutions. Fig.~\ref{fig:objspace} illustrates this for the same example used in Fig.~\ref{fig:searchspace}. As we can see, each cluster comprises bids with similar objective function values, and the best region includes the best candidate solutions although they are located in different areas of the search space. 

For updating based on clustering in objective space, the average (mid-point) of candidate solutions is used, in addition to the best solution from the winner cluster (i.e., from clustering in search space).

After clustering in both search and objective space, a new bid is updated as 
\begin{equation}
\label{eq:bidupd}
x_{i} = x_{i} + C_{1} r  (W - x_{i}) + C_{2}  r  (\bar{x} - x_{i}) ,
\end{equation}
where $\bar{x}$ is the centroid of the bids of the best cluster found in objective space clustering, $C_1$ and $C_2$ are constants, and $r$ is a random number in $[0,1]$ taken from the uniform distribution.

\begin{table}[b!]
	\centering
	\caption{Parameter settings.}
	\label{par_set}
	\begin{tabular}{llc}
		\hline
		algorithm & parameter & value \\
		\hline
		CMA-ES~\cite{CMA_main_paper} & $\lambda$ & 50 \\
		\hline
		PSO~\cite{CEC2005} & $C_1$ & 2 \\
		& $C_2$ & 2 \\
		& $w$ & 1 to 0 \\
		\hline
		GWO~\cite{GWO_Main_Paper} & no parameters \\
		\hline
		WOA~\cite{WOA_Main_Paper} & $b$ & 1 \\
		\hline
		SSA~\cite{SSA_Main_Paper} & no parameters \\
		\hline
		HMS~\cite{HMS_Main_Paper} & number of clusters & 5 \\
		& $C$ & 1 \\
		& $M_L$ & 2 \\
		& $M_H$ & 5 \\
		\hline
		GHMS-RCS~\cite{GHMS-RCS} & number of clusters & 5 \\
		& $C$ & 1 \\
		& $M_L$ & 2 \\
		& $M_H$ & 5 \\
		& clustering probability & 0.5 \\
		\hline
		HMS-OS & number of clusters in search space & 5 \\
		& number of clusters in objective space & 10 \\
		& $C_{1}$ & 1.5 \\
		& $C_{2}$ & 1.5 \\
		& $M_L$ & 2 \\
		& $M_H$ & 10 \\
		\hline
	\end{tabular}
\end{table}

\subsection{Adaptive Selection of Mental Search Processes}
To further improve the proposed algorithm, we select the number of mental searches in an adaptive manner~\cite{HMS-IS-OSK}, rather than randomly as in the original HMS algorithm where it is determined as a random number between the minimum $M_{L}$ and the maximum number of mental process $M_{H}$.

In HMS-OS, the number of mental searches is chosen proportionally to the bid's quality as specified by its objective function value. Consequently, the vicinity of better bids are explored more, thus improving the exploitation ability of the algorithm.

In particular,  the number of mental searches $NML_{i}$ for bid $x_{i}$ is chosen as 
\begin{equation}
\label{eq:novel}
NML_{i}=M_{L}+round \left(\frac{N_{pop}-rank(x_i)+1}{N_{pop}} (M_{H}-M_{L}) \right) ,
\end{equation}  
where $rank(x_i)$ is the rank of $x_i$ obtained by sorting all bids by objective function value in ascending order (for a minimisation problem). As a consequence, $M_H$ mental searches are conducted for the best bid, whereas $M_L$ searches are performed for the worst bid.

\section{Experimental results}
\label{Sec:Exp}
To verify the efficacy of HMS-OS, we evaluate it on the CEC2017 benchmark functions~\cite{CEC2017}, 30 functions with different characteristics including unimodal functions (F1 to F3), multi-modal functions (F4 to F10), hybrid multi-modal functions (F11 to F20) and composite functions (F21 to F30). We compare our proposed algorithm with standard HMS algorithm~\cite{HMS_Main_Paper} and GHMS-RCS~\cite{GHMS-RCS}, a recent HMS variant, as well as several state-of-the-art population-based algorithms, including covariance matrix adaptation evolution strategy (CMA-ES)~\cite{CMA_main_paper}, particle swarm optimisation (PSO)~\cite{PSO_Main_Paper02}, grey wolf optimiser (GWO)~\cite{GWO_Main_Paper}, whale optimisation algorithm (WOA)~\cite{WOA_Main_Paper}, moth flame optimisation algorithm (MFO)~\cite{MFO_Main_Paper}, and salp swarm algorithm (SSA)~\cite{SSA_Main_Paper}. 

In all experiments, the maximum number of function evaluations and population size are set to $3000 \times D$ and 50, respectively, where $D$ is the dimensionality of the search space. The employed parameters for the various algorithms are given in Table~\ref{par_set}. Each algorithm is run 25 times, and we report the average and standard deviation over these 25 runs.

The results for $D=50$ are given in Table~\ref{Obj_D50}. In all tables, as a performance measure, we report the difference between the (known) optimal function value and the result obtained by the algorithm. From Table~\ref{Obj_D50}, it is apparent that HMS-OS obtains the best result for 26 of the 30 functions, while also giving the lowest standard deviation (and thus, best robustness) for 25 problems. Overall, HMS-OS yields the lowest average rank (1.30), and by a wide margin, thus clearly outperforming all other methods.

\begin{table*}[]
	\scriptsize
	\centering
	\caption{Results for all algorithms and all functions for $D=50$. Best results are bolded. }
	\begin{tabular}{l|c|cccccc|cc|c}
		function & & CMA-ES & PSO & GWO & WOA & MFO & SSA & HMS & HMS-RCS & HMS-OS \\ 
		\hline
		F1 & avg. & 4.64E+10 & 2.57E+08 & 5.59E+09 & 4.96E+08 & 3.75E+10 & \textbf{7.49E+03} & 2.99E+08 & 4.44E+08 & 3.43E+04 \\
		& std.dev. & 2.36E+10 & 1.81E+07 & 2.22E+09 & 3.64E+08 & 1.45E+10 & \textbf{8.69E+03} & 5.21E+08 & 3.90E+08 & 4.13E+04 \\ 
		\hline
		F2 & avg. & 5.07E+79 & 2.66E+25 & 7.96E+54 & 1.25E+70 & 6.94E+76 & 6.61E+32 & 1.42E+53 & 2.20E+50 & \textbf{5.23E+21} \\
		& std.dev. & 2.53E+80 & 8.24E+25 & 3.83E+55 & 6.24E+70 & 3.46E+77 & 2.70E+33 & 5.97E+53 & 7.86E+50 & \textbf{2.42E+22} \\ 
		\hline
		F3 & avg. & 3.75E+05 & \textbf{9.48E+03} & 9.42E+04 & 1.85E+05 & 1.67E+05 & 5.45E+04 & 3.48E+04 & 9.39E+04 & 1.47E+05 \\
		& std.dev. & 5.24E+04 & \textbf{3.35E+03} & 1.78E+04 & 5.14E+04 & 9.30E+04 & 1.85E+04 & 9.04E+03 & 2.62E+04 & 1.91E+04 \\ 
		\hline
		F4 & avg. & 7.61E+03 & 1.62E+02 & 7.14E+02 & 5.30E+02 & 4.06E+03 & 1.94E+02 & 3.04E+02 & 3.61E+02 & \textbf{4.62E+01} \\
		& std.dev. & 2.12E+03 & 6.32E+01 & 5.45E+02 & 1.32E+02 & 4.43E+03 & 4.26E+01 & 5.47E+01 & 1.41E+02 & \textbf{2.05E+01} \\ 
		\hline
		F5 & avg. & 3.08E+02 & 3.94E+02 & 2.07E+02 & 4.70E+02 & 4.62E+02 & 3.31E+02 & 2.38E+02 & 4.56E+02 & \textbf{1.63E+02} \\
		& std.dev. & 2.78E+02 & 4.13E+01 & 2.98E+01 & 6.62E+01 & 8.55E+01 & 8.80E+01 & 3.67E+01 & 5.92E+01 & \textbf{1.10E+01} \\ 
		\hline
		F6 & avg. & 6.80E+01 & 6.95E+01 & 1.76E+01 & 8.30E+01 & 4.78E+01 & 5.60E+01 & 1.87E+01 & 3.76E+01 & \textbf{5.70E+00} \\
		& std.dev. & 7.09E+00 & 7.54E+00 & 4.16E+00 & 1.33E+01 & 8.21E+00 & 6.99E+00 & 7.55E+00 & 6.03E+00 & \textbf{8.30E-01} \\ 
		\hline
		F7 & avg. & 2.20E+02 & 5.01E+02 & 3.59E+02 & 1.06E+03 & 1.13E+03 & 4.08E+02 & 3.00E+02 & 4.44E+02 & \textbf{2.04E+02} \\
		& std.dev. & 1.53E+02 & 3.40E+01 & 4.42E+01 & 1.40E+02 & 4.26E+02 & 8.23E+01 & 4.66E+01 & 1.45E+02 & \textbf{1.20E+01} \\ 
		\hline
		F8 & avg. & 5.17E+02 & 4.15E+02 & 2.19E+02 & 4.49E+02 & 4.79E+02 & 3.14E+02 & 2.54E+02 & 4.43E+02 & \textbf{1.60E+02} \\
		& std.dev. & 1.94E+02 & 3.12E+01 & 3.42E+01 & 5.78E+01 & 8.61E+01 & 6.09E+01 & 4.26E+01 & 5.67E+01 & \textbf{1.34E+01} \\ 
		\hline
		F9 & avg. & 1.35E+04 & 2.47E+04 & 8.16E+03 & 2.76E+04 & 1.71E+04 & 1.22E+04 & 6.46E+03 & 2.12E+04 & \textbf{2.84E+03} \\
		& std.dev. & 1.02E+04 & 3.74E+03 & 4.45E+03 & 9.49E+03 & 5.07E+03 & 2.47E+03 & 1.87E+03 & 6.15E+03 & \textbf{6.75E+02 }\\ 
		\hline
		F10 & avg. & 1.32E+04 & 8.82E+03 & 6.50E+03 & 1.01E+04 & 7.47E+03 & 7.00E+03 & 7.53E+03 & 7.49E+03 & \textbf{4.28E+03} \\
		& std.dev. & 4.77E+02 & 6.89E+02 & 1.61E+03 & 1.46E+03 & 1.12E+03 & 7.97E+02 & 1.09E+03 & 8.04E+02 & \textbf{2.51E+02} \\ 
		\hline
		F11 & avg. & 6.45E+04 & 3.59E+02 & 2.97E+03 & 1.17E+03 & 1.28E+04 & 3.78E+02 & 8.25E+02 & 9.04E+02 & \textbf{1.65E+02} \\
		& std.dev. & 1.32E+04 & 7.65E+01 & 9.80E+02 & 3.42E+02 & 1.10E+04 & 9.95E+01 & 1.55E+02 & 2.35E+02 & \textbf{1.59E+01} \\ 
		\hline
		F12 & avg. & 2.24E+10 & 1.07E+08 & 5.44E+08 & 5.79E+08 & 4.62E+09 & 4.25E+07 & 1.07E+08 & 9.99E+07 & \textbf{4.75E+05} \\
		& std.dev. & 5.35E+09 & 2.99E+07 & 7.75E+08 & 3.77E+08 & 4.50E+09 & 2.66E+07 & 6.89E+07 & 4.71E+07 & \textbf{2.55E+05} \\ 
		\hline
		F13 & avg. & 1.29E+10 & 1.47E+07 & 1.05E+08 & 5.55E+06 & 8.80E+08 & 1.80E+05 & 1.17E+05 & 7.20E+04 & \textbf{2.67E+03} \\
		& std.dev. & 2.06E+09 & 2.93E+06 & 1.45E+08 & 5.72E+06 & 1.18E+09 & 1.54E+05 & 4.19E+04 & 3.11E+04 & \textbf{2.61E+03} \\ 
		\hline
		F14 & avg. & 2.17E+07 & 1.17E+05 & 9.45E+05 & 1.97E+06 & 1.08E+06 & 1.44E+05 & 2.85E+05 & 3.40E+05 & \textbf{2.12E+04} \\
		& std.dev. & 1.13E+07 & 7.10E+04 & 1.15E+06 & 1.69E+06 & 2.18E+06 & 1.25E+05 & 1.33E+05 & 1.95E+05 & \textbf{1.23E+04} \\ 
		\hline
		F15 & avg. & 2.51E+09 & 3.05E+06 & 1.56E+07 & 1.08E+06 & 8.65E+06 & 5.73E+04 & 3.24E+04 & 2.89E+04 & \textbf{5.68E+03} \\
		& std.dev. & 1.10E+09 & 1.11E+06 & 2.42E+07 & 2.50E+06 & 2.36E+07 & 3.54E+04 & 1.77E+04 & 1.47E+04 & \textbf{4.31E+03} \\ 
		\hline
		F16 & avg. & 5.37E+03 & 2.15E+03 & 1.46E+03 & 3.81E+03 & 2.73E+03 & 2.00E+03 & 2.10E+03 & 2.39E+03 & \textbf{1.35E+03} \\
		& std.dev. & 3.93E+02 & 5.46E+02 & 3.94E+02 & 9.35E+02 & 6.69E+02 & 5.07E+02 & 4.36E+02 & 4.36E+02 & \textbf{1.95E+02} \\ 
		\hline
		F17 & avg. & 1.05E+03 & 1.68E+03 & 1.06E+03 & 2.55E+03 & 2.30E+03 & 1.75E+03 & 1.51E+03 & 1.79E+03 & \textbf{8.55E+02} \\
		& std.dev. & 3.95E+02 & 3.41E+02 & 2.70E+02 & 6.15E+02 & 5.13E+02 & 4.31E+02 & 4.24E+02 & 3.98E+02 & \textbf{8.40E+01} \\ 
		\hline
		F18 & avg. & 1.27E+08 & 1.50E+06 & 5.10E+06 & 1.41E+07 & 2.96E+06 & 1.95E+06 & 1.78E+06 & 2.77E+06 & \textbf{2.04E+05} \\
		& std.dev. & 7.61E+07 & 8.78E+05 & 5.93E+06 & 8.68E+06 & 2.91E+06 & 1.36E+06 & 1.38E+06 & 1.84E+06 & \textbf{1.04E+05} \\ 
		\hline
		F19 & avg. & 1.10E+09 & 2.78E+06 & 3.16E+06 & 5.13E+06 & 5.93E+07 & 2.89E+06 & 3.12E+04 & 1.75E+04 & \textbf{1.64E+04} \\
		& std.dev. & 6.94E+08 & 1.52E+06 & 7.51E+06 & 4.96E+06 & 2.50E+08 & 1.50E+06 & 1.20E+04 & 1.77E+04 & \textbf{6.99E+03} \\ 
		\hline
		F20 & avg. & 1.62E+03 & 1.30E+03 & 8.97E+02 & 1.70E+03 & 1.40E+03 & 1.22E+03 & 1.10E+03 & 1.27E+03 & \textbf{7.86E+02} \\
		& std.dev. & 3.02E+02 & 2.86E+02 & 2.96E+02 & 3.28E+02 & 4.64E+02 & 3.00E+02 & 2.95E+02 & 2.14E+02 & \textbf{1.03E+02} \\ 
		\hline
		F21 & avg. & 7.99E+02 & 6.61E+02 & 3.94E+02 & 8.43E+02 & 6.41E+02 & 4.60E+02 & 4.91E+02 & 6.36E+02 & \textbf{3.65E+02} \\
		& std.dev. & 4.19E+01 & 6.97E+01 & 3.11E+01 & 9.68E+01 & 8.44E+01 & 5.07E+01 & 4.42E+01 & 4.52E+01 & \textbf{9.31E+00} \\ 
		\hline
		F22 & avg. & 1.43E+04 & 9.09E+03 & 6.78E+03 & 1.01E+04 & 8.28E+03 & 6.04E+03 & 7.85E+03 & 8.08E+03 & \textbf{4.40E+03} \\
		& std.dev. & \textbf{4.31E+02} & 2.02E+03 & 9.82E+02 & 1.37E+03 & 1.10E+03 & 2.40E+03 & 1.19E+03 & 9.93E+02 & 8.58E+02 \\ 
		\hline
		F23 & avg. & 1.14E+03 & 1.55E+03 & 6.64E+02 & 1.35E+03 & 8.41E+02 & 7.14E+02 & 8.14E+02 & 8.27E+02 & \textbf{6.11E+02} \\
		& std.dev. & 4.47E+01 & 2.96E+02 & 7.96E+01 & 1.66E+02 & 5.10E+01 & 8.52E+01 & 6.13E+01 & 4.48E+01 & 2\textbf{.16E+01} \\ 
		\hline
		F24 & avg. & 1.14E+03 & 1.21E+03 & \textbf{7.24E+02} & 1.30E+03 & 8.16E+02 & 7.48E+02 & 8.78E+02 & 8.98E+02 & 8.01E+02 \\
		& std.dev. & 4.87E+01 & 1.48E+02 & 6.75E+01 & 1.45E+02 & 5.77E+01 & 6.11E+01 & 3.92E+01 & 5.34E+01 & \textbf{1.51E+01} \\ 
		\hline
		F25 & avg. & 2.72E+03 & \textbf{5.12E+02} & 1.06E+03 & 9.11E+02 & 3.08E+03 & 5.69E+02 & 6.34E+02 & 7.54E+02 & 5.25E+02 \\
		& std.dev. & 1.42E+03 & 3.92E+01 & 2.29E+02 & 1.26E+02 & 2.61E+03 & 2.80E+01 & 7.33E+01 & 1.30E+02 & \textbf{1.82E+01} \\ 
		\hline
		F26 & avg. & 8.76E+03 & 6.09E+03 & 3.95E+03 & 1.05E+04 & 5.82E+03 & 3.36E+03 & 5.47E+03 & 5.32E+03 & \textbf{2.51E+03} \\
		& std.dev. & 5.70E+02 & 3.69E+03 & 5.47E+02 & 1.43E+03 & 5.32E+02 & 2.03E+03 & 5.90E+02 & \textbf{2.76E+02} & 1.25E+03 \\ 
		\hline
		F27 & avg. & 1.12E+03 & 1.07E+03 & 8.99E+02 & 1.69E+03 & 8.75E+02 & 8.10E+02 & 7.22E+02 & 7.40E+02 & \textbf{5.91E+02} \\
		& std.dev. & 7.24E+01 & 7.16E+02 & 1.08E+02 & 3.84E+02 & 1.04E+02 & 9.12E+01 & 8.10E+01 & 9.75E+01 & \textbf{2.17E+01} \\ 
		\hline
		F28 & avg. & 6.77E+03 & 4.97E+02 & 1.48E+03 & 1.20E+03 & 5.11E+03 & 5.23E+02 & 7.09E+02 & 2.67E+03 & \textbf{4.88E+02} \\
		& std.dev. & 1.77E+02 &\textbf{ 1.78E+01} & 4.74E+02 & 1.77E+02 & 1.01E+03 & 2.34E+01 & 5.42E+02 & 1.66E+03 & 1.84E+01 \\ 
		\hline
		F29 & avg. & 1.04E+04 & 2.55E+03 & 1.62E+03 & 4.85E+03 & 2.31E+03 & 2.42E+03 & 1.67E+03 & 1.54E+03 & \textbf{1.10E+03} \\
		& std.dev. & 3.06E+03 & 3.51E+02 & 3.13E+02 & 1.35E+03 & 5.63E+02 & 4.45E+02 & 3.42E+02 & 3.13E+02 & \textbf{1.13E+02} \\ 
		\hline
		F30 & avg. & 2.43E+09 & 5.58E+07 & 1.09E+08 & 1.35E+08 & 1.29E+08 & 7.09E+07 & 2.31E+06 & 2.22E+06 & \textbf{1.02E+06} \\ 
		& std.dev. & 8.57E+08 & 7.14E+06 & 3.28E+07 & 6.09E+07 & 3.26E+08 & 1.57E+07 & 1.25E+06 & 7.47E+05 & \textbf{1.54E+05} \\ 
		\hline
		\multicolumn{2}{l|}{average rank} & 7.87 & 5.02 & 4.43 & 7.60 & 6.90 & 3.60 & 3.65 & 4.63 & \textbf{1.30} \\ 
		\hline
		\multicolumn{2}{l|}{overall rank} & 9 & 6 & 4 & 8 & 7 & 2 & 3 & 5 & \textbf{1} \\ 
		\hline
	\end{tabular}
	\label{Obj_D50}
\end{table*}

For $D=100$, the results are given in Table~\ref{Obj_D100}, HMS-OS obtains the best results for all but one function, thus impressively demonstrating superiority over all other evaluated techniques.

\begin{table*}[]
	\scriptsize
	\centering
	\caption{Results for all algorithms and all functions for $D=100$.}
	\begin{tabular}{l|c|cccccc|cc|c}
		function & & CMA-ES & PSO & GWO & WOA & MFO & SSA & HMS & HMS-RCS & HMS-OS \\ 
		\hline
		F1 & avg. & 4.20E+10 & 1.07E+09 & 4.25E+10 & 1.98E+07 & 1.24E+11 & 6.49E+04 & 1.51E+10 & 1.14E+10 & \textbf{1.09E+04} \\
		& std.dev. & 5.82E+10 & 9.06E+07 & 7.90E+09 & 6.83E+06 & 3.65E+10 & 2.78E+05 & 6.50E+09 & 4.01E+09 & \textbf{8.99E+03} \\ 
		\hline
		F2 & avg. & 1.04E+166 & 8.45E+71 & 4.30E+133 & 1.15E+148 & 7.99E+159 & 7.41E+108 & 6.53E+134 & 8.24E+124 & \textbf{7.32E+48} \\
		& std.dev. & 1.76E+166 & 2.89E+72 & 2.15E+134 & 5.42E+148 & 6.65E+159 & 3.70E+109 & 3.27E+135 & 4.12E+125 & \textbf{3.60E+49} \\ 
		\hline
		F3 & avg. & 8.50E+05 & 2.36E+05 & 2.45E+05 & 7.55E+05 & 6.55E+05 & 3.24E+05 & 1.14E+05 & 2.78E+05 & 3.94E+05 \\
		& std.dev. & 8.76E+04 & 3.39E+04 & 1.99E+04 & 1.50E+05 & 1.90E+05 & 7.16E+04 & 1.37E+04 & 4.12E+04 & 2.81E+04 \\ 
		\hline
		F4 & avg. & 1.71E+04 & 3.87E+02 & 3.36E+03 & 6.69E+02 & 2.38E+04 & 4.07E+02 & 2.01E+03 & 1.53E+03 & \textbf{1.91E+02} \\
		& std.dev. & 3.03E+03 & 8.27E+01 & 9.04E+02 & 1.04E+02 & 1.68E+04 & 5.74E+01 & 6.83E+02 & 4.20E+02 & \textbf{2.47E+01} \\ 
		\hline
		F5 & avg. & 1.07E+03 & 1.09E+03 & 6.41E+02 & 9.25E+02 & 1.20E+03 & 8.55E+02 & 6.78E+02 & 1.32E+03 & \textbf{2.55E+02} \\
		& std.dev. & 3.53E+02 & 7.27E+01 & 1.23E+02 & 6.65E+01 & 1.79E+02 & 9.36E+01 & 7.06E+01 & 1.02E+02 & \textbf{1.87E+01} \\ 
		\hline
		F6 & avg. & 1.44E+01 & 8.60E+01 & 3.65E+01 & 7.93E+01 & 7.32E+01 & 6.47E+01 & 3.32E+01 & 6.33E+01 & \textbf{3.21E+00} \\
		& std.dev. & 2.94E+01 & 4.87E+00 & 4.12E+00 & 8.75E+00 & 6.51E+00 & 5.58E+00 & 6.76E+00 & 6.62E+00 & \textbf{4.76E-01} \\ 
		\hline
		F7 & avg. & 8.65E+02 & 1.25E+03 & 1.25E+03 & 2.57E+03 & 4.04E+03 & 1.22E+03 & 9.67E+02 & 1.44E+03 & \textbf{3.99E+02} \\
		& std.dev. & 8.55E+01 & 1.11E+02 & 1.61E+02 & 1.54E+02 & 7.69E+02 & 2.04E+02 & 1.41E+02 & 2.67E+02 & \textbf{2.28E+01} \\ 
		\hline
		F8 & avg. & 1.27E+03 & 1.20E+03 & 6.33E+02 & 1.12E+03 & 1.27E+03 & 8.85E+02 & 7.14E+02 & 1.35E+03 & \textbf{2.61E+02} \\
		& std.dev. & 5.40E+01 & 7.42E+01 & 5.36E+01 & 1.34E+02 & 1.86E+02 & 1.69E+02 & 8.10E+01 & 1.35E+02 & \textbf{1.53E+01} \\ 
		\hline
		F9 & avg. & 3.25E+04 & 6.32E+04 & 3.27E+04 & 3.54E+04 & 4.21E+04 & 2.71E+04 & 2.22E+04 & 5.52E+04 & \textbf{6.94E+03} \\
		& std.dev. & 2.19E+04 & 5.47E+03 & 1.20E+04 & 9.46E+03 & 5.96E+03 & 3.69E+03 & 3.71E+03 & 8.59E+03 & \textbf{2.05E+03} \\ 
		\hline
		F10 & avg. & 3.04E+04 & 2.20E+04 & 1.57E+04 & 1.95E+04 & 1.68E+04 & 1.48E+04 & 1.79E+04 & 1.86E+04 & \textbf{1.09E+04} \\
		& std.dev. & 5.62E+02 & 1.26E+03 & 3.15E+03 & 2.43E+03 & 1.98E+03 & 1.63E+03 & 2.28E+03 & 1.50E+03 & \textbf{7.67E+02} \\ 
		\hline
		F11 & avg. & 4.51E+05 & 2.44E+03 & 5.33E+04 & 1.17E+04 & 1.32E+05 & 5.04E+03 & 1.35E+04 & 2.56E+04 & \textbf{9.98E+02} \\
		& std.dev. & 8.10E+04 & \textbf{2.05E+02} & 1.20E+04 & 6.65E+03 & 8.92E+04 & 1.40E+03 & 5.70E+03 & 8.92E+03 & \textbf{2.05E+02} \\ 
		\hline
		F12 & avg. & 5.27E+10 & 8.27E+08 & 7.40E+09 & 7.23E+08 & 3.48E+10 & 4.48E+08 & 1.52E+09 & 1.26E+09 & \textbf{8.89E+05} \\
		& std.dev. & 8.63E+09 & 2.05E+08 & 4.63E+09 & 2.62E+08 & 1.81E+10 & 2.12E+08 & 9.57E+08 & 4.70E+08 & \textbf{2.94E+05} \\ 
		\hline
		F13 & avg. & 1.16E+10 & 5.08E+07 & 4.15E+08 & 6.20E+04 & 4.43E+09 & 9.40E+04 & 3.78E+07 & 1.63E+07 & \textbf{3.09E+03} \\
		& std.dev. & 2.27E+09 & 6.15E+06 & 4.20E+08 & 2.74E+04 & 4.07E+09 & 4.42E+04 & 8.85E+07 & 4.13E+07 & \textbf{3.10E+03} \\ 
		\hline
		F14 & avg. & 1.15E+08 & 1.57E+06 & 4.95E+06 & 1.45E+06 & 8.14E+06 & 1.63E+06 & 2.07E+06 & 3.73E+06 & \textbf{1.60E+05} \\
		& std.dev. & 3.94E+07 & 3.72E+05 & 3.30E+06 & 6.26E+05 & 7.46E+06 & 7.35E+05 & 1.85E+06 & 1.91E+06 & \textbf{6.22E+04} \\ 
		\hline
		F15 & avg. & 5.73E+09 & 1.52E+07 & 1.02E+08 & 6.21E+04 & 1.81E+09 & 6.54E+04 & 5.00E+05 & 1.79E+05 & \textbf{2.22E+03} \\
		& std.dev. & 1.10E+09 & 2.99E+06 & 1.26E+08 & 2.44E+04 & 1.48E+09 & 2.32E+04 & 1.79E+06 & 2.52E+05 & \textbf{2.23E+03} \\ 
		\hline
		F16 & avg. & 1.20E+04 & 5.85E+03 & 4.42E+03 & 8.43E+03 & 6.36E+03 & 5.05E+03 & 6.06E+03 & 5.78E+03 & \textbf{2.63E+03} \\
		& std.dev. & 6.59E+02 & 8.16E+02 & 6.28E+02 & 1.83E+03 & 8.90E+02 & 6.80E+02 & 8.21E+02 & 8.86E+02 & \textbf{2.50E+02} \\ 
		\hline
		F17 & avg. & 4.22E+04 & 4.32E+03 & 3.54E+03 & 5.47E+03 & 6.44E+03 & 3.60E+03 & 5.25E+03 & 5.04E+03 & \textbf{2.02E+03} \\
		& std.dev. & 2.36E+04 & 5.77E+02 & 1.97E+03 & 8.08E+02 & 1.73E+03 & 5.59E+02 & 6.43E+02 & 6.64E+02 & \textbf{2.20E+02} \\ 
		\hline
		F18 & avg. & 1.39E+08 & 3.18E+06 & 4.83E+06 & 1.88E+06 & 1.63E+07 & 3.03E+06 & 5.67E+06 & 7.38E+06 & \textbf{3.74E+05} \\
		& std.dev. & 3.92E+07 & 1.17E+06 & 2.90E+06 & 6.42E+05 & 2.06E+07 & 1.53E+06 & 4.72E+06 & 4.72E+06 & \textbf{1.16E+05} \\ 
		\hline
		F19 & avg. & 4.44E+09 & 2.67E+07 & 1.15E+08 & 1.33E+07 & 7.69E+08 & 1.21E+07 & 1.54E+06 & 4.28E+05 & \textbf{1.34E+03} \\
		& std.dev. & 1.30E+09 & 7.05E+06 & 1.50E+08 & 6.89E+06 & 1.19E+09 & 5.16E+06 & 2.76E+06 & 4.07E+05 & \textbf{1.64E+03} \\ 
		\hline
		F20 & avg. & 5.14E+03 & 3.77E+03 & 3.28E+03 & 4.23E+03 & 3.78E+03 & 3.10E+03 & 3.65E+03 & 3.71E+03 & \textbf{2.02E+03} \\
		& std.dev. & 3.66E+02 & 5.29E+02 & 1.08E+03 & 5.17E+02 & 5.67E+02 & 5.01E+02 & 5.90E+02 & 5.16E+02 & \textbf{2.46E+02} \\ 
		\hline
		F21 & avg. & 1.44E+03 & 1.69E+03 & 8.69E+02 & 1.79E+03 & 1.55E+03 & 1.12E+03 & 1.10E+03 & 1.61E+03 & \textbf{4.88E+02} \\
		& std.dev. & 3.00E+02 & 1.48E+02 & 5.41E+01 & 2.49E+02 & 1.50E+02 & 1.90E+02 & 8.34E+01 & 9.62E+01 & \textbf{1.62E+01} \\ 
		\hline
		F22 & avg. & 3.11E+04 & 2.44E+04 & 1.74E+04 & 2.17E+04 & 1.84E+04 & 1.68E+04 & 1.99E+04 & 1.94E+04 & \textbf{1.21E+04} \\
		& std.dev. & 8.83E+02 & 1.19E+03 & 3.14E+03 & 2.04E+03 & 1.79E+03 & 1.68E+03 & 1.84E+03 & 1.26E+03 & \textbf{7.15E+02} \\ 
		\hline
		F23 & avg. & 1.84E+03 & 2.91E+03 & 1.23E+03 & 2.44E+03 & 1.49E+03 & 1.29E+03 & 1.41E+03 & 1.32E+03 & \textbf{7.45E+02} \\
		& std.dev. & 5.95E+01 & 3.89E+02 & 9.33E+01 & 2.59E+02 & 1.14E+02 & 1.42E+02 & 7.82E+01 & 5.61E+01 & \textbf{1.60E+01} \\ 
		\hline
		F24 & avg. & 2.44E+03 & 3.04E+03 & 1.75E+03 & 3.55E+03 & 1.94E+03 & 1.66E+03 & 1.95E+03 & 1.88E+03 & \textbf{1.16E+03} \\
		& std.dev. & 9.19E+01 & 2.79E+02 & 9.40E+01 & 3.51E+02 & 1.33E+02 & 1.39E+02 & 1.28E+02 & 9.99E+01 & \textbf{1.47E+01} \\ 
		\hline
		F25 & avg. & 8.27E+03 & 9.45E+02 & 3.42E+03 & 1.13E+03 & 1.12E+04 & 9.97E+02 & 1.59E+03 & 2.87E+03 & \textbf{7.29E+02} \\
		& std.dev. & 4.32E+03 & 4.88E+01 & 6.58E+02 & 9.39E+01 & 7.94E+03 & 6.60E+01 & 3.23E+02 & 7.11E+02 & \textbf{2.88E+01} \\ 
		\hline
		F26 & avg. & 1.93E+04 & 1.71E+04 & 1.22E+04 & 2.82E+04 & 1.53E+04 & 1.23E+04 & 1.58E+04 & 1.46E+04 & \textbf{6.17E+03} \\
		& std.dev. & 6.74E+02 & 7.38E+03 & 1.31E+03 & 3.53E+03 & 1.40E+03 & 2.57E+03 & 9.16E+02 & 7.49E+02 & \textbf{1.80E+02} \\ 
		\hline
		F27 & avg. & 2.00E+03 & 6.43E+02 & 1.33E+03 & 2.27E+03 & 1.28E+03 & 1.14E+03 & 8.75E+02 & 8.36E+02 & \textbf{6.00E+02} \\
		& std.dev. & 1.78E+02 & 9.87E+01 & 1.22E+02 & 7.89E+02 & 2.39E+02 & 1.56E+02 & 7.88E+01 & 8.44E+01 & \textbf{8.77E+00} \\ 
		\hline
		F28 & avg. & 1.91E+04 & 6.38E+02 & 4.73E+03 & 9.13E+02 & 1.62E+04 & 7.35E+02 & 2.37E+03 & 1.30E+04 & \textbf{5.80E+02} \\
		& std.dev. & 8.78E+02 & 7.62E+01 & 1.02E+03 & 6.08E+01 & 1.59E+03 & 4.41E+01 & 1.04E+03 & 1.67E+02 & \textbf{2.40E+01} \\ 
		\hline
		F29 & avg. & 1.34E+04 & 6.94E+03 & 5.28E+03 & 1.11E+04 & 8.22E+03 & 6.39E+03 & 5.04E+03 & 4.82E+03 & \textbf{2.72E+03} \\
		& std.dev. & 1.19E+04 & 5.16E+02 & 6.24E+02 & 2.62E+03 & 4.36E+03 & 9.99E+02 & 6.85E+02 & 6.21E+02 & \textbf{1.70E+02} \\ 
		\hline
		F30 & avg. & 1.02E+10 & 1.07E+08 & 1.03E+09 & 1.83E+08 & 2.33E+09 & 9.79E+07 & 1.02E+07 & 1.99E+06 & \textbf{1.43E+04} \\
		& std.dev. & 2.09E+09 & 2.75E+07 & 1.03E+09 & 7.47E+07 & 1.73E+09 & 3.68E+07 & 9.33E+06 & 1.51E+06 & \textbf{5.61E+03} \\ 
		\hline
		\multicolumn{2}{l|}{average rank} & 7.75 & 5.25 & 4.72 & 5.90 & 7.25 & 3.33 & 4.50 & 5.13 & \textbf{1.17} \\ 
		\hline
		\multicolumn{2}{l|}{overall rank} & 9 & 6 & 4 & 7 & 8 & 2 & 3 & 5 & \textbf{1} \\ 
		\hline
	\end{tabular}
	\label{Obj_D100}
\end{table*}

We also perform a Wilcoxon signed-rank test between HMS-OS and the other algorithms whose results are given in Table~\ref{Tab:wilx}. As we can see from there, the obtained $p$-value is below 0.05 in all cases, confirming that HMS-OS gives statistically significantly better results than all other methods.

\begin{table}[t!]
	\centering
	\caption{Results ($p$-values) of Wilcoxon signed rank test between HMS-OS and other algorithms.}
	\begin{tabular}{l|cc}
		\hline
		& $D=50$ & $D=100$ \\ 
		\hline
		HMS vs.\ CMA-ES & 1.7333 E-06 & 2.6966E-11 \\
		HMS vs.\ PSO & 1.7988 E-05 & 1.8839E-10 \\
		HMS vs.\ GWO & 2.3704E-05 & 7.5409E-10 \\
		HMS vs.\ WOA & 1.7343E-06 & 1.6295E-11 \\
		HMS vs.\ MFO & 1.7343E-06 & 1.6295E-11 \\
		HMS vs.\ SSA & 1.4772E-04 & 1.5821E-10 \\
		HMS vs.\ HMS & 1.6394E-05 & 8.6652E-10 \\
		HMS vs.\ HMS-RCS & 1.6394E-05 & 1.3042E-10 \\ 
		\hline
	\end{tabular}
	\label{Tab:wilx}
\end{table}

\begin{figure}[t!]
	\centering
	\includegraphics[width=.38\textwidth]{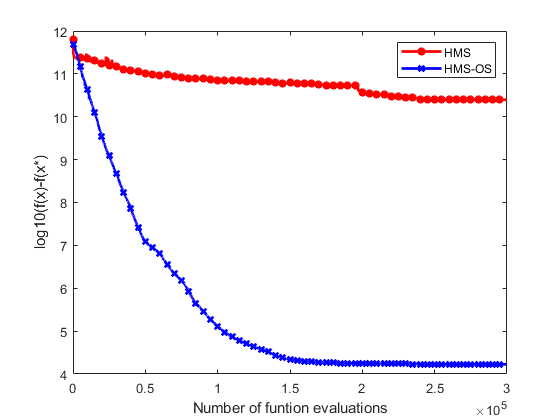} 
	\includegraphics[width=.38\textwidth]{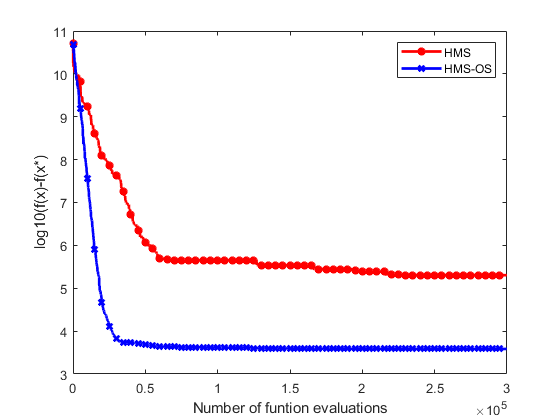}
	\includegraphics[width=.38\textwidth]{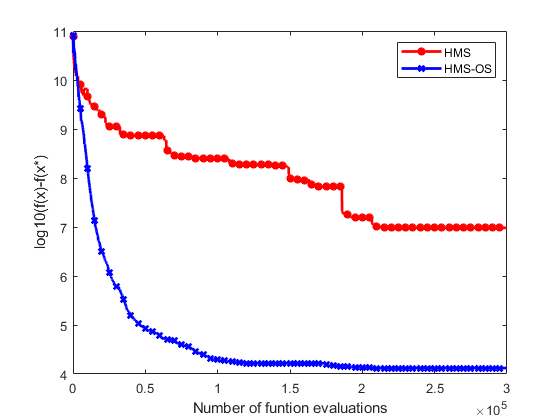}
	\caption{Convergence plots for F1 (top), F15 (middle), and F30 (bottom).}
	\label{fig:Convergence_Curve}
\end{figure}

Fig.~\ref{fig:Convergence_Curve} shows convergence curves of our proposed algorithm compared to those of the original HMS algorithm, for three representative functions (F1, F15, and F30).  As we can observe, HMS-OS consistently converges faster than standard HMS.

As our HMS-OS algorithm includes two main modifications, we investigate the effect of each modification separately. The results for $D=100$, as a representative, are given in Table~\ref{tab:components} where HMS-OS-V1 refers to HMS with (only) adaptive selection of mental search processes, while HMS-OS-V2 denotes HMS with (only) our proposed clustering algorithm. It can be seen that in most cases HMS-OS-V2 yields the second lowest objective function value, while HMS-OS-V1 is ranked third in most cases, showing that the effect of clustering also in objective space  on the overall performance is higher than the adaptive mental process selection.

\begin{table}[b!]
	\caption{Results for the each of the proposed modifications.}
	\begin{tabular}{lcccc}
		\hline
		& HMS       & HMS-OS-V1 & HMS-OS-V2 & HMS-OS   \\ \hline
		F1           & 1.51E+10  & 1.17E+07  & 2.29E+04  & 1.09E+04 \\
		F2           & 6.53E+134 & 2.89E+48  & 3.23E+56  & 7.32E+48 \\
		F3           & 1.14E+05  & 3.58E+05  & 4.10E+05  & 3.94E+05 \\
		F4           & 2.01E+03  & 4.75E+02  & 2.13E+02  & 1.91E+02 \\
		F5           & 6.78E+02  & 7.17E+02  & 2.59E+02  & 2.55E+02 \\
		F6           & 3.32E+01  & 4.83E+01  & 8.44E+00  & 3.21E+00 \\
		F7           & 9.67E+02  & 9.99E+02  & 4.50E+02  & 3.99E+02 \\
		F8           & 7.14E+02  & 6.84E+02  & 2.58E+02  & 2.61E+02 \\
		F9           & 2.22E+04  & 3.55E+04  & 1.08E+04  & 6.94E+03 \\
		F10          & 1.79E+04  & 1.15E+04  & 1.32E+04  & 1.09E+04 \\
		F11          & 1.35E+04  & 1.23E+03  & 3.21E+03  & 9.98E+02 \\
		F12          & 1.52E+09  & 3.47E+06  & 1.75E+06  & 8.89E+05 \\
		F13          & 3.78E+07  & 8.65E+03  & 4.22E+03  & 3.09E+03 \\
		F14          & 2.07E+06  & 2.42E+05  & 2.35E+05  & 1.60E+05 \\
		F15          & 5.00E+05  & 3.96E+03  & 2.44E+03  & 2.22E+03 \\
		F16          & 6.06E+03  & 4.17E+03  & 3.00E+03  & 2.63E+03 \\
		F17          & 5.25E+03  & 2.76E+03  & 2.26E+03  & 2.02E+03 \\
		F18          & 5.67E+06  & 5.83E+05  & 6.27E+05  & 3.74E+05 \\
		F19          & 1.54E+06  & 2.40E+03  & 2.07E+03  & 1.34E+03 \\
		F20          & 3.65E+03  & 2.20E+03  & 2.30E+03  & 2.02E+03 \\
		F21          & 1.10E+03  & 8.08E+02  & 4.88E+02  & 4.88E+02 \\
		F22          & 1.99E+04  & 1.27E+04  & 1.40E+04  & 1.21E+04 \\
		F23          & 1.41E+03  & 1.01E+03  & 7.36E+02  & 7.45E+02 \\
		F24          & 1.95E+03  & 1.42E+03  & 1.13E+03  & 1.16E+03 \\
		F25          & 1.59E+03  & 1.10E+03  & 7.70E+02  & 7.29E+02 \\
		F26          & 1.58E+04  & 9.78E+03  & 5.84E+03  & 6.17E+03 \\
		F27          & 8.75E+02  & 7.20E+02  & 6.11E+02  & 6.00E+02 \\
		F28          & 2.37E+03  & 1.15E+03  & 5.95E+02  & 5.80E+02 \\
		F29          & 5.04E+03  & 3.50E+03  & 2.85E+03  & 2.72E+03 \\
		F30          & 1.02E+07  & 4.55E+04  & 2.38E+04  & 1.43E+04 \\ \hline
		average rank &   3.77        & 2.87          & 2.11          & 1.25         \\ \hline
		overall rank &   4        &   3        &    2       &  1        \\ \hline
	\end{tabular}
	\label{tab:components}
\end{table}

\section{Conclusions}
\label{Sec:conc}
In this paper, we have proposed a novel human mental search (HMS) algorithm, HMS-OS, to solve global optimisation problems. While in standard HMS, grouping is performed in search space, in HMS-OS the population is clustered both in search and objective space. The new updating scheme then employs both the best candidate solution of the best region selected by search space clustering and the centroid of the members of the best region selected by objective space clustering. In addition, the number of mental searches is adaptively selected so that the vicinity of the better candidate solutions is explored more. Extensive experiments on the CEC-2017 benchmark functions for dimensionalities of 50 and 100, convincingly confirm HMS-OS to outperform HMS and a number of other population-based metaheuristic algorithms including covariance matrix adaptation evolution strategy, particle swarm optimisation, grey wolf optimiser, whale optimisation algorithm, moth flame optimisation algorithm, and salp swarm algorithm.

\bibliographystyle{IEEEtran}
\bibliography{jalal}
\end{document}